\documentclass[twocolumn,letterpaper]{article}
\usepackage[]{aaai23}  % DO NOT CHANGE THIS
\usepackage{times}  % DO NOT CHANGE THIS
\usepackage{helvet}  % DO NOT CHANGE THIS
\usepackage{courier}  % DO NOT CHANGE THIS
\usepackage[hyphens]{url}  % DO NOT CHANGE THIS
\usepackage{graphicx} % DO NOT CHANGE THIS
\usepackage{natbib}  % DO NOT CHANGE THIS AND DO NOT ADD ANY OPTIONS TO IT
\usepackage{caption} % DO NOT CHANGE THIS AND DO NOT ADD ANY OPTIONS TO IT
\usepackage{threeparttable}
\usepackage{amssymb}
\usepackage{multirow}
\usepackage{subfigure}
\usepackage[colorlinks,linkcolor=red]{hyperref}
\usepackage{algorithm}
\usepackage{algorithmic}
\usepackage{newfloat}
\usepackage{listings}
\usepackage{booktabs}
\DeclareCaptionStyle{ruled}{labelfont=normalfont,labelsep=colon,strut=off} % DO NOT CHANGE THIS
\floatstyle{ruled}
\newfloat{listing}{tb}{lst}{}
\floatname{listing}{Listing}
\nocopyright
\title{BEVDepth: Acquisition of Reliable Depth for Multi-view 3D Object Detection}
\author{
    Yinhao Li\textsuperscript{\rm 1, \rm2},
    Zheng Ge\textsuperscript{\rm 3},
    Guanyi Yu\textsuperscript{\rm 3},
    Jinrong Yang\textsuperscript{\rm 4}\\
    Zengran Wang\textsuperscript{\rm 3},
    Yukang Shi\textsuperscript{\rm 5},
    Jianjian Sun\textsuperscript{\rm 3},
    Zeming Li\textsuperscript{\rm 3}
}
\affiliations{
    \textsuperscript{\rm 1}Key Laboratory of Intelligent Information Processing, Institute of Computing Technology, CAS,\\
    \textsuperscript{\rm 2}University of Chinese Academy of Sciences,
    \textsuperscript{\rm 3}MEGVII Technology,\\
    \textsuperscript{\rm 4}Huazhong University of Science and Technology
    \textsuperscript{\rm 5}Xi’an Jiaotong University\\
    liyinhao20@mails.ucas.edu.cn,\\
    \{gezheng, yuguanyi, yangjinrong, wangzengran, shiyukang, sunjianjian, lizeming\}@megvii.com
}

\usepackage{bibentry}
\begin{document}

\maketitle

\begin{abstract}

In this research, we propose a new 3D object detector with a trustworthy depth estimation,  dubbed BEVDepth, for camera-based Bird's-Eye-View~(BEV) 3D object detection. Our work is based on a key observation -- depth estimation in recent approaches is surprisingly inadequate given the fact that depth is essential to camera 3D detection. Our BEVDepth resolves this by leveraging explicit depth supervision. A camera-awareness depth estimation module is also introduced to facilitate the depth predicting capability. Besides, we design a novel Depth Refinement Module to counter the side effects carried by imprecise feature unprojection. Aided by customized Efficient Voxel Pooling and multi-frame mechanism, BEVDepth achieves the new state-of-the-art 60.9\% NDS on the challenging nuScenes test set while maintaining high efficiency. For the first time, the NDS score of a camera model reaches 60\%. Code is released at \href{https://github.com/Megvii-BaseDetection/BEVDepth}{https://github.com/Megvii-BaseDetection/BEVDepth}.

\end{abstract}

\begin{figure*}[!t]
\centering
\includegraphics[width=\textwidth]{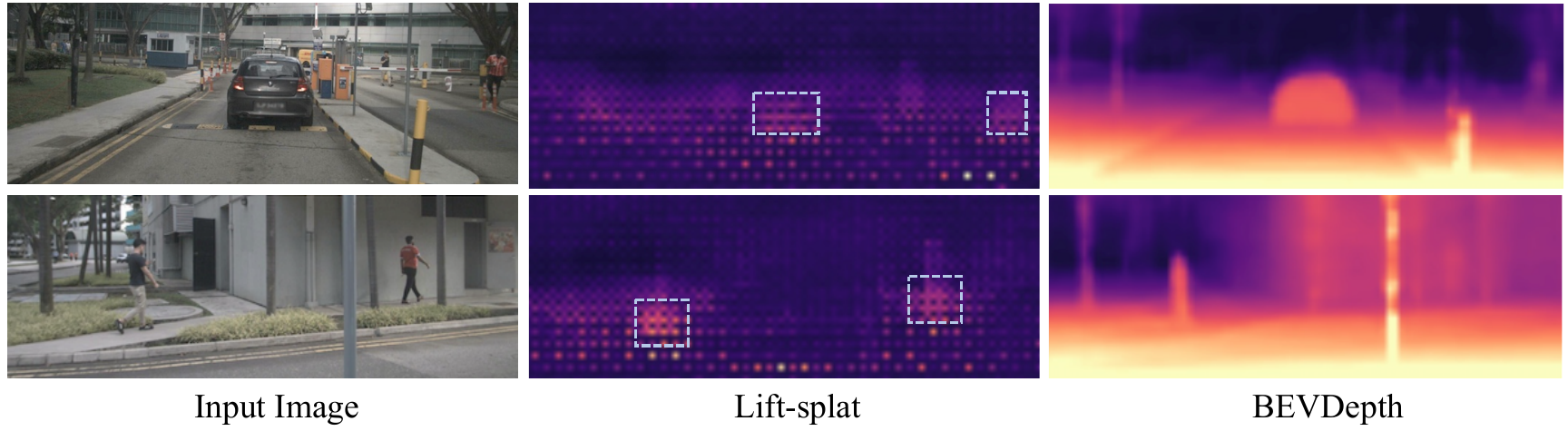}
\caption{Depth estimation results in Lift-splat detector and BEVDepth. Dashed boxes highlight the regions that Lift-splat detector makes ``relatively'' accurate depth predictions in, usually being the attaching regions between objects and the ground.}\label{teaser}
\end{figure*}

\section{Introduction} \label{sec:intro}

LiDAR and camera are the two main sensors used by the current autonomous systems to detect 3D objects and perceive the environment. While LiDAR-based methods have demonstrated their ability to deliver trustworthy 3D detection results, multi-view camera-based methods have recently attracted increasing attention because of their lower cost.

The feasibility of using multi-view cameras for 3D perception has been well addressed in LSS~\cite{philion2020lift}. They first ``lift'' multi-view features to 3D frustums using estimated depth, then ``splat'' frustums onto a reference plane, usually being a plane in Bird's-Eye-View (BEV). The BEV representation is non-trivial since it not only enables an end-to-end training scheme of a multiple input cameras system but also provides a unified space for various downstream tasks such as BEV segmentation, object detection~\cite{huang2021bevdet, li2022bevformer} and motion planning. However, despite the success of LSS-based perception algorithms, the learned depth within this pipeline is barely studied. We ask -- \emph{does the quality of learned depth within these detectors really meet the requirement for precise 3D object detection?}

We attempt to answer this question qualitatively first by visualizing the estimated depth (Fig.~\ref{teaser}) in a Lift-splat based detector. Even though the detector achieves 30 mAP on nuScenes~\cite{caesar2020nuscenes} benchmark, its depth are surprisingly poor. Only a few region of features predict reasonable depth and contribute to subsequent tasks (see dashed boxes in Fig.~\ref{teaser}), while most other regions do not. Based on this observation, we point out that the depth learning mechanism in existing Lift-splat brings three deficiencies:

\begin{itemize}
\item \textbf{Inaccurate Depth} Since the depth prediction module is indirectly supervised by the final detection loss, the absolute depth quality is far from satisfying;
\item \textbf{Depth Module Over-fitting} Most pixels can not predict reasonable depth, meaning that they are not properly trained during the learning stage. It makes us doubt about depth module's generalizing ability.
\item \textbf{Imprecise BEV Semantics} The learned depth in Lift-splat unprojects image features into 3D frustum features, which will be further pooled into BEV features. With a poor depth like in Lift-splat, only part of features are unprojected to correct BEV positions, resulting in imprecise BEV semantics.
\end{itemize}

\noindent We will dive deep into these three deficiencies in Sec.~\ref{sec3}. 

Moreover, we reveal the great potential of improving depth by replacing the learned depth in Lift-splat with its ground-truth generated from point cloud data. As a result, mAP and NDS are both boosted by nearly 20\%. The translation error (mATE) decreases as well, from 0.768 to 0.393. Such a phenomenon clearly reveals that enhancing depth is the key to high-performance camera 3D detection.

Therefore, in this work, we introduce BEVDepth, a new multi-view 3D detector that leverages depth supervision derives from point clouds to guide depth learning. We are the first team that presents a thorough analysis of how the depth quality affects the overall system. Meanwhile, we innovatively propose to encode camera intrinsics and extrinsics into a depth learning module so that the detector is robust to various camera settings. In the end, a Depth Refinement Module is further introduced to refine the learned depth. 

To validate the power of BEVDepth, we test it on nuScenes~\cite{caesar2020nuscenes} dataset -- a well-known benchmark in the field of 3D detection. Aided by our customized Efficient Voxel Pooling and Multi-frame Fusion technique, BEVDepth achieves 60.9\% NDS on the nuScenes \emph{test} set, being the new state-of-the-art on this challenging benchmark while still maintaining high efficiency.

\section{Related Work} \label{rworks}
\subsection{Vision-based 3D object detection}
The goal of vision-based 3D detection is to predict the 3D bounding boxes of objects. It is an ill-posed problem because estimating the depth of objects from monocular images is inherently ambiguous. Even when multi-view cameras are available, estimating depth in areas without overlapping views remains challenging. Therefore, depth modeling is a critical component of vision-based 3D detection. One branch of research predicts 3D bounding boxes directly from 2D image features. 2D detectors, such as CenterNet ~\cite{zhou2019objects}, can be used for 3D detection with minor changes to detection heads. M3D-RPN ~\cite{brazil2019m3d} proposes depth-aware convolutional layers to enhance spatial awareness. D$^4$LCN ~\cite{huo2020learning} employs depth maps to guide dynamic kernel learning. By converting 3D targets into the image domain, FCOS3D ~\cite{wang2021fcos3d} predicts 2D and 3D attributes of objects. Further, PGD ~\cite{wang2022probabilistic} presents geometric relation graphs to facilitate depth estimation for 3D object detection. DD3D ~\cite{park2021pseudo} demonstrates that depth pre-training can significantly improve end-to-end 3D detection.

Another line of work predicts objects in 3D space. There are many ways to convert 2D image features into 3D space. One typical approach is transforming image-based depth maps to pseudo-LiDAR to mimic the LiDAR signal ~\cite{wang2019pseudo, you2019pseudo, qian2020end}. Image features can also be used to generate 3D voxels ~\cite{rukhovich2022imvoxelnet} or orthographic feature maps ~\cite{roddick2018orthographic}. LSS ~\cite{philion2020lift} proposes a view transform method that explicitly predicts depth distribution and projects image features onto a bird's-eye view (BEV), which has been proved practical for 3D object detection ~\cite{reading2021categorical, huang2021bevdet, huang2022bevdet4d}. BEVFormer ~\cite{li2022bevformer} performs 2D-to-3D transformation with local attention and grid-shaped BEV queries. Following DETR ~\cite{carion2020end}, DETR3D ~\cite{wang2022detr3d} detects 3D objects with transformers and object queries, and PETR ~\cite{liu2022petr} improves performance further by introducing 3D position-aware representations.

\subsection{LiDAR-based 3D object detection}

Due to the accuracy of depth estimation, LiDAR-based 3D detection methods are frequently employed in autonomous driving perception tasks. VoxelNet~\cite{zhou2018voxelnet} voxelizes the point cloud, converting it from sparse to dense voxels, and then proposes bounding boxes in dense space to aid the index during convolution.
SECOND~\cite{yan2018second} increases performance on the KITTI dataset~\cite{geiger2012we} by introducing a more effective structure and gt-sampling technique based on VoxelNet~\cite{zhou2018voxelnet}. Sparse convolution is also used in SECOND~\cite{yan2018second} to boost speed. PointPillars~\cite{lang2019pointpillars} encodes point clouds using pillars rather than 3D convolution processes, making it fast but maintaining good performance.
CenterPoint~\cite{yin2021center} proposes an anchor-free detector that extends CenterNet~\cite{zhou2019objects} to 3D space and achieves high performance on nuScenes dataset~\cite{caesar2020nuscenes} and Waymo open dataset~\cite{sun2020scalability}. PointRCNN~\cite{shi2019pointrcnn}, unlike the grid-based approaches discussed above, creates proposals directly from point clouds. It then employs LiDAR segmentation to identify foreground points for proposals and produce bounding boxes in the second stage. \cite{qi2019deep,yang2022dbq} use Hough voting to collect point features and then propose bounding boxes from clusters. Because of its dense feature representation, grid-based approaches are faster, but they lose information from raw point clouds, whereas point-based methods can connect raw point clouds but are inefficient when locating neighbors for each point. PV-RCNN~\cite{shi2020pv} is proposed to preserve efficiency while allowing adjustable receptive fields for point features.

\subsection{Depth Estimation}
Depth prediction is critical for monocular image interpretation. Fu et al.~\cite{fu2018deep} employ a regression method to predict the depth of an image using dilated convolution and a scene understanding module. Monodepth~\cite{godard2017unsupervised} predicts depth without supervision using disparity and reconstruction. Monodepth2~\cite{godard2019digging} uses a combination of depth estimation and pose estimation networks to forecast depth in a single frame.

Some approaches predict depth by constructing cost-volume. MVSNet~\cite{yao2018mvsnet} first introduces cost-volume to the field of depth estimation. Based on MVSNet, RMVSNet~\cite{yao2019recurrent} uses GRU to reduce memory cost, MVSCRF~\cite{xue2019mvscrf} adds CRF module, Cascade MVSNet~\cite{gu2020cascade} changes MVSNet to cascade structure. Wang et al.~\cite{wang2021patchmatchnet} generate depth prediction using multi-scale fusion and introduce adaptive modules which improve performance and reduce memory consumption at the same time. Bae et al.~\cite{bae2022multi} fuse single-view images with multi-view images and introduce depth-sampling to reduce the cost of computation.

\section{Delving into Depth Prediction in Lift-splat} \label{sec3}

In Sec.~\ref{sec:intro}, we show that a LSS-based detector with surprisingly poor depth can still obtain reasonable 3D detection results. In this section, we first review the overall structure of our baseline 3D detector built on Lift-splat. Then we conduct a simple experiment on our base detector to reveal why we observe the previous phenomenon. Finally, we discuss three deficiencies carried by this detector and point out a potential solution to it.

\begin{table}[!t]
\centering
% \vspace{-2mm}
% \resizebox{\textwidth}{!}{
\begin{tabular}{c|ccc} 
\toprule
\textbf{$D^{pred}$} &\textbf{mAP}$\uparrow$   & \textbf{mATE}$\downarrow$ & \textbf{NDS}$\uparrow$ \\
\midrule
 learned    & 0.282 & 0.768 & 0.327      \\
 random soft& 0.245 & 0.838 & 0.290   \\
 random hard& 0.176 &	0.922 & 0.224 \\ \midrule
 ground truth & 0.470 & 0.393 & 0.515 \\
 \bottomrule
\end{tabular}
\caption{Evaluation of depth prediction on the nuScenes \emph{val} set. ``soft'' and ``hard'' denote gaussian and one-hot randomization along depth dimension, respectively.}\label{tab:random}
\end{table}

\subsection{Model Architecture for Base Detector} Our vanilla Lift-splat based detector simply replaces the segmentation head in LSS~\cite{philion2020lift} with CenterPoint~\cite{yin2021center} head for 3D detection. Specifically, it consists of four main components shown in Fig.~\ref{fig:bevdepth}. 1) An Image Encoder (\emph{e.g.}, ResNet~\cite{resnet}) that extracts 2D features $F^{2d} = \{ F_i^{2d} \in \mathbb{R}^{C_F \times H \times W}, i=1,2,...,N\}$ from $N$ view input images $I = \{I_i, i=1,2,...,N\}$, where $H$, $W$ and $C_F$ stand for feature's height, width and channel number; 2) A DepthNet that estimates images depth $D^{pred} = \{ D_i^{pred} \in \mathbb{R}^{C_D \times H \times W}, i=1,2,...,N\}$ from image features $F^{2d}$, where $C_D$ stands for the number of depth bins; 3) A View Transformer that projects $F^{2d}$ in 3D representations $F^{3d}$ using Eq.~\ref{eq1} then pools them into an integrated BEV representation $F^{bev}$; 4) A 3D Detection Head predicting the class, 3D box offset and other attributes.

\begin{equation}\label{eq1}
    F_i^{3d} = F_i^{2d} \otimes D_i^{pred}, \quad F_i^{3d} \in \mathbb{R}^{C_F \times C_D \times H  \times W}.
\end{equation}

\subsection{Making Lift-splat work is easy} 

The learned depth $D^{pred}$ is believed essential since it is used to build the BEV representation for subsequent tasks. However, the poor visualization results in Fig.~\ref{teaser} contradict this consensus. In Sec.~\ref{sec:intro}, we attribute the success of Lift-splat to partially reasonable learned depth. Now, we take a step further to study the essence of this pipeline by replacing $D^{pred}$ with a random initialized tensor and freezing it during both the training and testing phases. Results are shown in Table~\ref{tab:random}. We are surprised to find that mAP only drops 3.7\% (from 28.2\% to 24.5\%) after replacing $D^{pred}$ with randomized soft values. We hypothesize that even if the depth used for unprojecting features is catastrophically broken, the soft nature of depth distribution still helps unproject to the right depth position to some extent, and thus obtains a reasonable mAP, nevertheless it simultaneously unprojects much non-negligible
noise. We further replace the soft randomized depth with a hard randomized depth (one-hot activation at each position) and observe a greater drop by 6.9\%, verifying our assumption. This demonstrates that as long as the depth at the correct position has activation, the detection head can work. It also explains why the learned depth is poor in most areas in Fig.~\ref{teaser}, but the detection mAP is still reasonable.

\begin{table}[!t]
\centering
\begin{tabular}{c|c|cccc}
\toprule
\textbf{Region}       & \textbf{DL} & \textbf{SILog}$\downarrow$ & \textbf{AbsRel}$\downarrow$ & \textbf{SqRel} & \textbf{RMSE}$\downarrow$  \\ \midrule
\multirow{2}{*}{All}  &    & 54.58 & 3.03   & 85.11 & 19.45 \\
                      &  \checkmark  & 27.62 & 0.23   & 2.09  & 5.78  \\ \midrule
\multirow{2}{*}{Best} &    & 27.87 & 0.38   & 6.96  & 8.29  \\
                      &  \checkmark  & 14.12 & 0.10    & 1.04  & 4.55  \\ \bottomrule
\end{tabular}
\caption{Evaluation of depth prediction on the nuScenes \emph{val} set. DL denotes Depth Loss. All foreground points are taken for evaluation.}\label{depth_eval}
\end{table}

\subsection{Making Lift-splat work well is hard}
Although obtaining reasonable results, the existing performance is far from satisfying. In this part, we reveal three deficiencies in the existing working mechanism of Lift-splat, including inaccurate depth, depth module over-fitting and imprecise BEV semantics. To demonstrate our idea more clearly, we compare two baselines -- one is the naive LSS-based detector, named Base Detector, and another one utilizes extra depth supervision derives from the point clouds data on $D^{pred}$, which will be described in detail in Sec.~\ref{sec4}. We name it Enhanced Detector.

\paragraph{Inaccurate depth} In Base Detector, the gradients on the depth module derives from the detection loss, which is indirect. It is natural to study the quality of learned depth. Therefore, We evaluate the learned depth $D^{pred}$ on nuScenes \emph{val} using the commonly used depth estimation metric~\cite{deptheval} including scale invariant logarithmic error (SILog), mean absolute relative error (Abs Rel), mean squared relative error (Sq Rel) and root mean squared error (RMSE). We evaluate two detectors under two different protocols: 1) all pixels for each object and 2) the best-predicted pixel for each object. Results are shown in Table~\ref{depth_eval}. When evaluating all foreground regions, the Base Detector only achieves 3.03 AbsRel, which is greatly poor than existing depth estimation algorithms~\cite{li2022depthformer,  bhat2021adabins}. However, as for Enhanced Detector, the AbsRel is largely reduced from 3.03 to 0.23, which becomes a more reasonable value. It is worth mentioning that performance of Base Detector under the best matching protocol is almost comparable to the Enhanced Detector under all-region protocol. This verifies our assumption in Sec.~\ref{sec:intro} that when a detector is trained without depth loss (just like Lift-splat), it detects objects by only learning partial depth. After applying depth loss on the best matching protocol, the learned depth is further improved. All of these results demonstrate that the implicitly learned depth is inaccurate and is far from satisfying.

\begin{figure}[!t]
\includegraphics[width=0.45\textwidth]{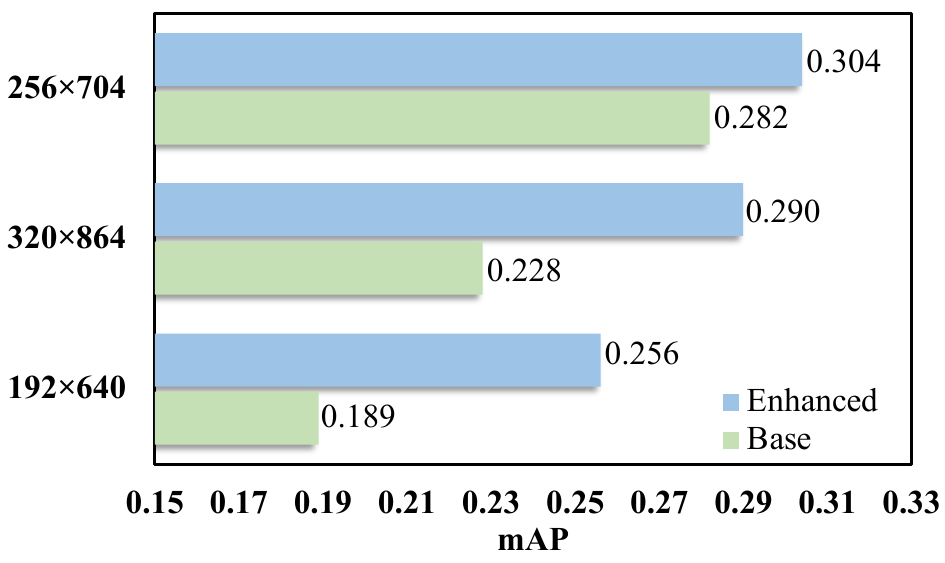}
\caption{Testing detectors' robustness to image sizes. We use $256\times704$ for training. mAP on nuScenes are reported.}
\vspace{-.1in}
\label{sizes}
\end{figure}

\paragraph{Depth Module Over-fitting} As we stated in the previous content, the Base Detector only learns to predict depth in partial regions. Most pixels are not trained to predict reasonable depth, which raises our concern about the depth module's generalizing ability. Concretely, the detector learning depth in that way could be very sensitive to hyper-parameters such as image sizes, camera parameters, \emph{etc}. To verify this, we choose ``image size'' as the variable, and conduct the following experiment to study the model's generalizing ability: we first train the Base Detector and the Enhanced Detector using input size 256$\times$704. Then we test them using 192$\times$640, 256$\times$704 and 320$\times$864 sizes, respectively. As we can see in Fig.~\ref{sizes}, the Base Detector loses more accuracy when testing image size is inconsistent with the training image size. The performance loss for Enhance Detector is much less. Such a phenomenon implies that the model without depth loss has a higher risk of over-fitting, and thus it may also be sensitive to the noise in camera intrinsics, extrinsics, or other hyper-parameters. 

\begin{figure}[!t]
\includegraphics[width=0.48\textwidth]{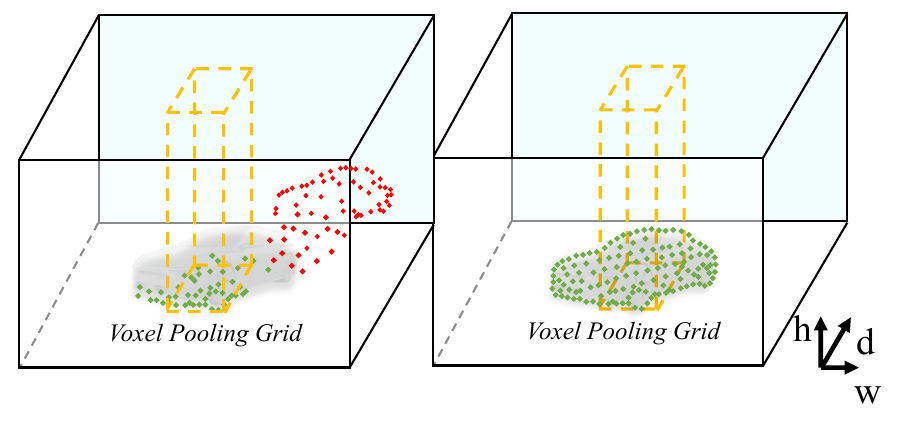}
\caption{Compared to the Base Detector (left), the Enhanced Detector (right) retains more structure information during feature unprojection and thus can provide precise semantics. Each dot denotes an image feature.}
\label{unproj}
\end{figure}

\paragraph{Imprecise BEV Semantics} Once image features are unprojected to frustum features using learned depth, a Voxel/Pillar Pooling operation is adopted to aggregate them to BEV. Fig.~\ref{unproj} shows that image features are not properly unprojected without depth supervision. Therefore, the pooling operation only aggregates part of semantic information. The Enhanced Detector performs better in this scenario. We hypothesize that the poor depth is harmful to the classification task. Then we use the classification heatmaps from both models and evaluate their TP / (TP + FN) as an indicator for comparison, where a TP represents an anchor point/feature which is assigned as the positive sample and is correctly classified by the CenterPoint head while FN represents the opposite meaning. See Table~\ref{tab:cls-acc}, the Enhanced Detector consistently outperforms the other one under different positive thresholds, which verifies our assumption.

Driven by the above analysis, we realize the necessity of endowing a better depth in multi-view 3D detectors, and propose our solution to it -- BEVDepth.

\begin{table}
\centering
\begin{tabular}{c|ccc} 
\toprule
\textbf{Method} &\textbf{th=0.3}   & \textbf{th=0.5} & \textbf{th=0.7} \\
\midrule
 Base Detector & 42.28\% & 18.36\% & 5.12\%     \\
 Enhanced Detector & 45.23\% & 22.47\% & 8.20\%    \\
\bottomrule
\end{tabular}
\caption{Classification on the nuScenes \emph{val} set. We use the classification heatmap for evaluation, \emph{th} denotes the threshold of heatmap.}
\label{tab:cls-acc}
\end{table}

\begin{figure*}[t]
\includegraphics[width=0.85\textwidth]{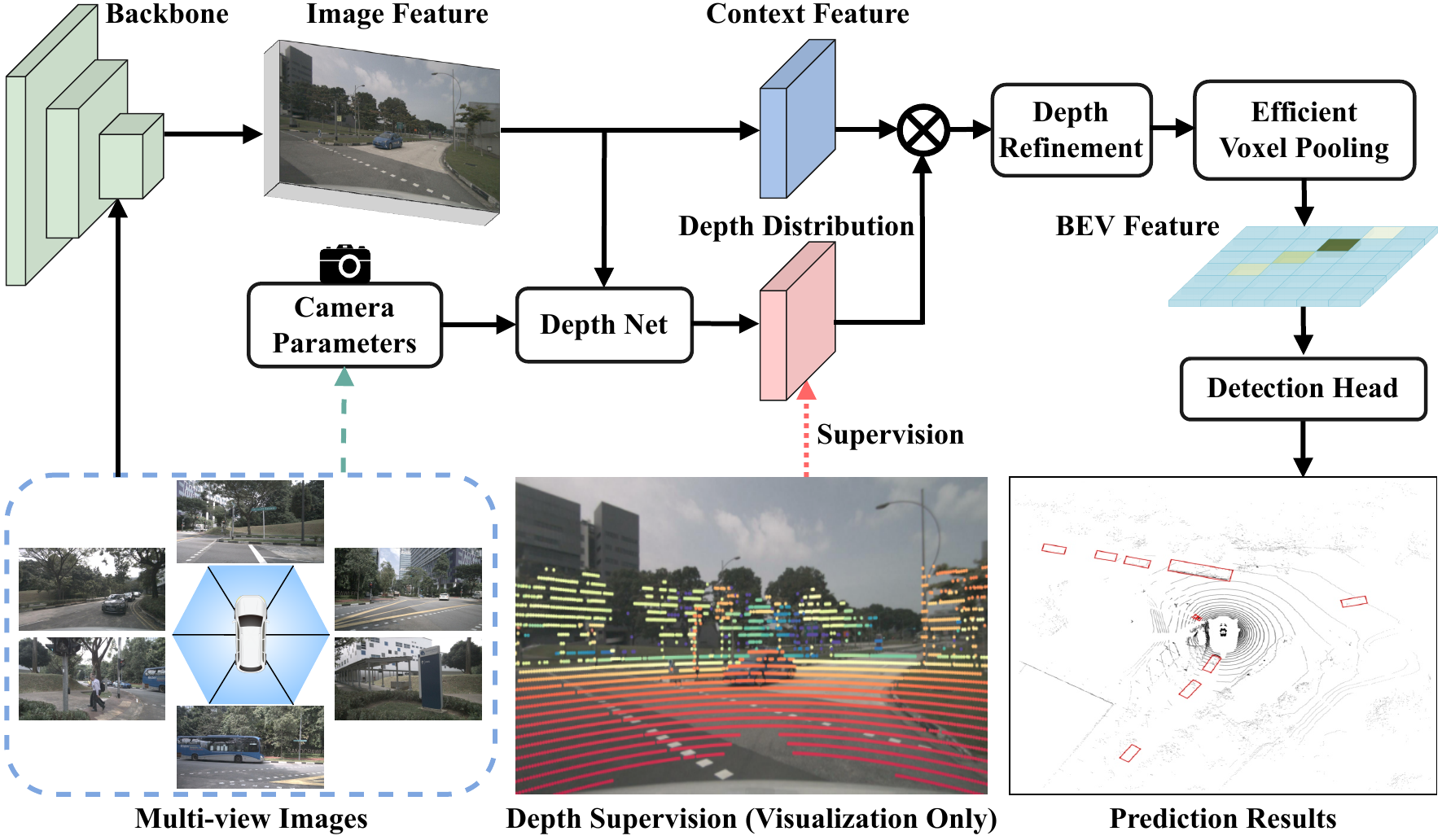}
\centering
\caption{Framework of BEVDepth. Image backbone extracts image feature from multi-view images. Depth net takes Image feature as input, generates context and depth, and gets the final point feature. Voxel Pooling unifies all point features into one coordinate system and pools them onto the BEV feature map.}
\label{fig:bevdepth}
\end{figure*}

\section{BEVDepth}\label{sec4}

BEVDepth is a new multi-view 3D detector with reliable depth. It leverages Explicit Depth Supervision on a Camera-aware Depth Prediction Module (DepthNet) with a novel Depth Refinement Module on unprojected frustum features to achieve this.

\paragraph{Explicit Depth Supervision} In Base Detector, the only supervision of the depth module comes from the detection loss. However, due to the difficulty of monocular depth estimation, a sole detection loss is far from enough to supervise the depth module. Therefore, we propose to supervise the intermediate depth prediction $D^{pred}$ using ground-truth $D^{gt}$ derived from point clouds data $P$. Denote $R_i \in \mathbb{R}^{3\times3}$ and $t_i \in \mathbb{R}^3$ as the rotation and translation matrix from the LiDAR coordinate to the camera coordinate of the $i^{th}$ view, and denote $K_i\in \mathbb{R}^{3\times3}$ as the $i^{th}$ camera's intrinsic parameter. To obtain $D^{gt}$, we first calculate:

\begin{equation}
    \hat{P_i^{img}}(ud, vd, d) = K_{i}(R_i P + t_i),
\end{equation}

\noindent which can be further converted to 2.5D image coordinates $P_i^{img}(u, v, d)$, where $u$ and $v$ denote coordinates in pixel coordinate. If the 2.5D projection of a certain point cloud does not fall into the $i^{th}$ view, we simply discard it. See Fig.~\ref{fig:bevdepth} for an example of the projection result. Then, to align the shape between the projected point clouds and the predicted depth, a \textit{min pooling} and a \textit{one hot} are adopted on $P_i^{img}$. We jointly define these two operations as $\phi$, the resulting $D^{gt}$ can thus be written in Eq.~\ref{dgt}. As for the depth loss $L_{depth}$, we simply adopt Binary Cross Entropy. 

\begin{equation}\label{dgt}
    D_i^{gt} = \phi(P_i^{img}).
\end{equation}

\paragraph{Camera-aware Depth Prediction}

According to the classic Camera Model, estimating depth is associated with the camera intrinsics, implying that it is non-trivial to model the camera intrinsics into DepthNet. This is especially important in multi-view 3D datasets when cameras may have different FOVs (\emph{e.g}., nuScenes Dataset). Therefore, we propose to utilize the camera intrinsics as one of the inputs for DepthNet. Concretly, the dimension for camera intrinsics is first scaled up to the features using an MLP layer. Then, they are used to re-weight the image feature $F_i^{2d}$ with an Squeeze-and-Excitation~\cite{senet} module. Finally, we concatenate the camera extrinsics to its intrinsics to help DepthNet aware of $F^{2d}$'s spatial location in the ego coordinate system. Denote $\psi$ as the original DepthNet, the overall Camera-awareness depth prediction can be written in:

\begin{equation}
    D_i^{pred} = \psi(SE(F_i^{2d}| MLP(\xi(R_i) \oplus \xi(t_i) \oplus \xi(K_i)))),
\end{equation}

\noindent where $\xi$ denotes the Flatten operation. An existing work ~\cite{dd3d} also leverages camera-awareness. They scale the regression targets according to cameras' intrinsics, making their method hard to adapt to automated systems with complex camera setups. Our method, on the other hand, models the cameras' parameters inside of the DepthNet, aiming at improving the intermediate depths' quality. Benefiting from the decoupled nature of LSS~\cite{philion2020lift}, the camera-aware depth prediction module is isolated from the detection head and thus the regression target, in this case, does not need to be changed, resulting in greater extensibility.

\paragraph{Depth Refinement Module} To further enhance the depth quality, we design a novel Depth Refinement Module. Specifically, we first reshape $F^{3d}$ from $[C_F, C_D, H, W]$ to $[C_F\times H, C_D, W]$, and stack several 3$\times$3 convolution layer on the $C_D \times W$ plane. Its output is finally reshaped back and fed into the subsequent Voxel/Pillar Pooling operation. On one hand, the Depth Refinement Module can aggregate features along the depth axis while the depth prediction confidence is low. One the other hand, when the depth prediction is inaccurate, the Depth Refinement Module is able to refine it to the correct position theoretically, as long as the receptive field is large enough. In a word, the Depth Refinement Module endows a rectification mechanism to the View Transformer stage, making it able to refine those improperly placed features.

\section{Experiment}

In this section, we first introduce our experimental setups. Then, comprehensive experiments are conducted on BEVDepth to validate the effects of our proposed components. Comparisons with other leading camera 3D detection models are presented in the end.

\begin{table}[!b]
\centering
% \vspace{-2mm}
\resizebox{0.475\textwidth}{!}{
\begin{tabular}{cccc|cccc}
\toprule
 \textbf{DL} & \textbf{CA} & \textbf{DR} &\textbf{MF} & \textbf{mAP}$\uparrow$  & \textbf{mATE}$\downarrow$ & \textbf{mAOE}$\downarrow$ & \textbf{NDS}$\uparrow$ \\
\midrule
    &   &     & & 0.282 &	0.768 & 0.698 & 0.327  \\
 \checkmark &   & & & 0.304 & 0.747 & 0.671 & 0.344  \\ 
 \checkmark & \checkmark & & & 0.314 & \textbf0.706 & 0.647 & 0.357    \\ 
 \checkmark & \checkmark & \checkmark & & \textbf0.322 & 0.707 & \textbf0.636 & \textbf0.367   \\ 
 \checkmark & \checkmark & \checkmark & \checkmark & \textbf{0.330} & {0.699} & \textbf{0.545} & \textbf{0.442}   \\ 
\bottomrule
\end{tabular}}
\caption{Ablation study of Depth Loss, Camera-awareness and Depth Refinement Module on the nuScenes \emph{val} set. DL, CA, DR and MF denotes Depth Loss, Camera-awareness, Depth Refinement Module and multi-frame, respectively.}
\label{tab:ablation-detection}
\end{table}

\begin{table}[!b]
\centering
% \vspace{-2mm}
% \resizebox{\textwidth}{!}{
\begin{tabular}{cc|cccc}
\toprule
 \textbf{BCE} & \textbf{L1}  & \textbf{mAP}$\uparrow$  & \textbf{mATE}$\downarrow$ & \textbf{mAOE}$\downarrow$ & \textbf{NDS}$\uparrow$ \\
\midrule
 \checkmark   & & 0.322 & 0.707 & 0.636 & 0.367   \\ 
 & \checkmark & 0.321 & \textbf{0.703} & 0.629 & 0.371   \\ 
 \checkmark & \checkmark  & \textbf{0.323} & 0.706 & \textbf{0.608} & \textbf{0.372}   \\ 
\bottomrule
\end{tabular}% }
\caption{Ablation study of different Depth Loss, including BCELoss and L1Loss. Results are reported on nuScenes \emph{val}.}
\label{tab:dloss}
\end{table}

\subsection{Experimental Setup}

\paragraph{Dataset and Metrics} nuScenes~\cite{caesar2020nuscenes} dataset is a large-scale autonomous driving benchmark containing data from six cameras, one LiDAR, and five radars. There are 1000 scenarios in the dataset, which are divided into 700, 150, and 150 scenes for training, validation, and testing, respectively. For 3D detection task, we report nuScenes Detection Score (NDS), mean Average Precision (mAP), as well as five True Positive (TP) metrics including mean Average Translation Error (mATE), mean Average Scale Error (mASE), mean Average Orientation Error (mAOE), mean Average Velocity Error (mAVE), mean Average Attribute Error (mAAE).

\paragraph{Implementation Details} Unless otherwise specified, we use ResNet-50~\cite{resnet} as the image backbone and the image size is processed to 256$\times$704. Following~\cite{huang2021bevdet}, we adopt image data augmentations including random cropping, random scaling, random flipping, and random rotation, and also adopt BEV data augmentations including random scaling, random flipping, and random rotation. We use AdamW~\cite{adamw} as an optimizer with a learning rate set to 2e-4 and batch size set to 64. For the ablation study, all experiments are trained for 24 epochs without using CBGS strategy~\cite{zhu2019class}. When compared to other methods, BEVDepth is trained for 20 epochs with CBGS. Camera-aware DepthNet is placed at the feature level with stride 16.

\subsection{Ablation Study}\label{sec:4.2}

\paragraph{Component Analysis} As shown in Table~\ref{tab:ablation-detection}, our vanilla BEVDepth achieves 28.2\% mAP and 32.7\% NDS. Adding Depth Loss improves mAP by 2.2\% which is consistent with our analysis -- Depth Loss is beneficial to classification. mATE marginally reduces 0.21, since the naive BEVDepth already learns to predict depth partially with the help of detection loss. Modeling camera parameters into DepthNet further reduces mATE by 0.41, revealing the importance of camera awareness. In the end, Depth Refinement Module improves 0.8\% mAP. We hypothesize that Depth Refinement Module makes features along the depth axis more compact, and thus is beneficial to reducing false response. Overall, our BEVDepth improves 4.0\% mAP and 4.0\% NDS compared to its baseline, showing the effectiveness of our innovations.

\begin{table}[!t]
\centering
% \vspace{-2mm}
% \resizebox{\textwidth}{!}{
\begin{tabular}{c|cccc}
\toprule
 \textbf{$C_D \times W$}   & \textbf{mAP}$\uparrow$  & \textbf{mATE}$\downarrow$ & \textbf{mAOE}$\downarrow$ & \textbf{NDS}$\uparrow$ \\
\midrule
- & 0.314 & 0.706 & 0.647 & 0.357   \\ 
1$\times$3 & 0.315 & 0.703 & 0.650 & 0.357   \\ 
3$\times$1 & 0.320 & \textbf{0.695}  & \textbf{0.624} & \textbf{0.369}  \\ 
3$\times$3 & \textbf{0.322} & 0.707 & 0.636 & 0.367   \\ 
\bottomrule
\end{tabular}% }\
\caption{Ablation study on the convolution kernel in Depth Refinement Module. Results are reported on nuScenes \emph{val}.}
\label{tab:kernel}
\vspace{-.1in}
\end{table}

\begin{table}[!h]
\centering
\resizebox{0.475\textwidth}{!}{
\begin{tabular}{l|c|cccccc|c} 
\toprule
\textbf{Method}             & \textbf{Resolution}   & \textbf{mAP}$\uparrow$ & \textbf{NDS}$\uparrow$ \\
\midrule
FCOS3D             & 900$\times$1600            & 0.295  & 0.372  \\
DETR3D             & 900$\times$1600            & 0.303 & 0.374  \\
BEVDet-R50         & 256$\times$704             & 0.286 & 0.372  \\
BEVDet-Tiny        & 512$\times$1408            & 0.349   & 0.417  \\
PETR-R50-DCN           & 384$\times$1056            & 0.313   & 0.381  \\
PETR-R101-DCN          & 512$\times$1408            & 0.357  & 0.421  \\
PETR-Tiny          & 512$\times$1408            & 0.361 & 0.431  \\
BEVDet4D-Tiny      & 256$\times$704             & 0.323 & 0.453  \\
BEVDet4D-Base      & 640$\times$1600            & 0.390 & 0.515  \\
BEVFormer-S        &       -                    & 0.375 & 0.448  \\
BEVFormer-R101-DCN & 900$\times$1600            & 0.416 & 0.517  \\
\midrule
BEVDepth-R50       & 256$\times$704             & 0.351 & 0.475  \\
BEVDepth-R101      & 512$\times$1408            & 0.412 & 0.535  \\
BEVDepth-R101-DCN      & 512$\times$1408            & \textbf{0.418} & \textbf{0.538}  \\
\bottomrule
\end{tabular}}
\caption{Comparison on the nuScenes \emph{val} set.}
\label{tab:val}
\vspace{-.1in}
\end{table}

\begin{table*}[!t]
\centering
% \vspace{-2mm}
\resizebox{0.95\textwidth}{!}{
\begin{tabular}{l|c|cccccc|c}
\toprule
\textbf{Method}                                   & \textbf{Modality} & \textbf{mAP}$\uparrow$  & \textbf{mATE}$\downarrow$ & \textbf{mASE}$\downarrow$  & \textbf{mAOE}$\downarrow$ & \textbf{mAVE}$\downarrow$ & \textbf{mAAE}$\downarrow$ & \textbf{NDS}$\uparrow$ \\
\midrule
CenterPoint                         & L        & 0.564 & - & - & -   & - & - & 0.648  \\ \midrule
FCOS3D~\cite{wang2021fcos3d}                                   & C        & 0.358 & 0.690 & 0.249 & 0.452   & 1.434 & 0.124 & 0.428  \\
DETR3D~\cite{wang2022detr3d}                                   & C        & 0.412 & 0.641 & 0.255 & 0.394   & 0.845 & 0.133 & 0.479  \\
BEVDet-Pure~\cite{huang2021bevdet}                              & C        & 0.398 & 0.556 & 0.239 & 0.414   & 1.010 & 0.153 & 0.463  \\
BEVDet-Beta                              & C        & 0.422 & 0.529 & 0.236 & 0.396   & 0.979 & 0.152 & 0.482  \\
PETR~\cite{liu2022petr} & C        & 0.434 & 0.641 & 0.248 & 0.437   & 0.894 & 0.143 & 0.481  \\
PETR-e                                   & C        & 0.441 & 0.593 & 0.249 & 0.384   & 0.808 & 0.132 & 0.504  \\
BEVDet4D~\cite{huang2022bevdet4d}                                 & C        & 0.451 & 0.511 & 0.241 & 0.386   & 0.301 & 0.121 & 0.569  \\
BEVFormer~\cite{li2022bevformer}                                & C        & 0.481 & 0.582 & 0.256 & 0.375   & 0.378 & 0.126 & 0.569  \\ 
PETRv2~\cite{liu2022petrv2}                                & C        & 0.490 & 0.561 & 0.243 & 0.361   & 0.343 & 0.120 & 0.582  \\ 
\midrule

BEVDepth                                 & C        & 0.503 & 0.445 & 0.245 & 0.378 & 0.320 & 0.126 & 0.600 \\

BEVDepth\dag                                & C        & 0.520 & 0.445 & 0.243 & 0.352 & 0.347 & 0.127 & \textbf{0.609} \\
\bottomrule
\end{tabular}
}
\caption{Comparison on the nuScenes \emph{test} set. L denotes LiDAR and C denotes camera. BEVDepth uses pretrained VovNet as backbone. the resolution of the input image is set to 640 $\times$ 1600. ~BEVDepth\dag  uses ConvNeXT~\cite{liu2022convnet} as backbone.}
\label{tab:test}
% \vspace{-.1in}
\end{table*}

\paragraph{Depth Loss} In the field of depth estimation, BCE and L1Loss are two common losses. In this part, we ablate the effect of using these two different losses in DepthNet (see Table~\ref{tab:dloss}), and find that different depth losses barely affect the final detection performance.

\paragraph{Depth Refinement Module} In Sec.~\ref{sec3}, we mention that Depth Refinement Module is designed to refine unsatisfactory depth by aggregating/refining the unprojected features along the depth axis. In terms of efficiency, we originally adopt 3$\times$3 convolution in it. Here we ablate different kernels including 1$\times$3, 3$\times$1 and 3$\times$3 to study its mechanism. See Table~\ref{tab:kernel}, when we use 1$\times$3 conv on $C_D \times W$ dimension, the information does not exchange along the depth axis, and the detection performance is barely affected. When we use 3$\times$1 conv, features are allowed to interact along the depth axis, mAP and NDS are correspondingly improved. This is similar to using naive $3\times 3$ conv, which reveals the nature of this module.

\subsection{Benchmark Results}

Here we briefly introduce two extra implementations that are crucial to obtain our performance on the nuScenes leardboard, \emph{i.e.}, Efficient Voxel Pooling and Multi-frame Fusion.

\paragraph{Efficient Voxel Pooling} Existing Voxel Pooling in Lift-splat leverages a ``cumsum trick'' that involves a ``sorting'' and a ``cumulative sum'' operations. Both operations are computationally inefficient. We propose to utilize great parallelism of GPU by assigning each frustum feature a CUDA thread that is used to add the feature to its corresponding BEV grid. As a result, the training time of our state-of-the-art model is reduced from 5 days to 1.5 days. The sole pooling operation is 80$\times$ faster than its baseline in Lift-splat.

\paragraph{Multi-frame Fusion} Multi-frame Fusion helps better detect objects and endows model ability to estimate velocity. We align the coordinates of frustum features from different frames into the current ego coordinate system to eliminate the effect of ego-motion and then perform Voxel Pooling. The pooled BEV features from different frames are directly concatenated and fed to following tasks.

\paragraph{nuScenes val set}We compare the proposed BEVDepth with other state-of-the-art methods like FCOS3D, DETR3D, BEVDet, PETR, BEVDet4D and BEVFormer on nuScenes \emph{val} set. We don't adopt test time augmentation. As can be seen from Table~\ref{tab:val}, BEVDepth shows superior performance in NDS (a key metric of nuScenes dataset), which improves 2\% over 2nd place, respectively. BEVDepth is also comparable with BEVFormer in mAP given the fact that they use stronger backbone and larger resolution input images. Using 256$\times$704 resolution input images, BEVDepth exceeds BEVDet on ResNet-50 by 10\% in NDS. BEVDepth also exceeds BEVDet4D-Tiny and BEVFormer-S by 2\% in NDS. When using 512$\times$1408 resolution input images, BEVDepth exceeds PETR on ResNet-101 6\% in mAP and 11\% in NDS. BEVDepth also exceeds BEVDET4D-Base 2\% in mAP and 2\% in NDS although their backbones are usually better than us.

% \vspace{-2mm}
\paragraph{nuScenes test set} For the submitted results on the \emph{test} set, we use the \emph{train} set and \emph{val} set for training. The result we submitted is a single model with test time augmentation. As listed in Table~\ref{tab:test}, BEVDepth ranks first on the nuScenes camera 3D objection leaderboard with a score of 50.3\% mAP and 60.0\% NDS. On mAP, we outperform the 2nd method PETRv2 by 1.3\%. On mATE, a key metric reflecting depth localization accuracy which is closely correlated to depth, we outperform PETRv2 by 11.6\%. On NDS, we surpass the second place by 1.8\%, and on other metrics, we remain at or on par with the best methods of the past. When switching the backbone to ConvNeXT, BEVDepth reaches 60.9\% NDS without extra data.

\section{Conclusion}

In this paper, a novel network architecture, namely BEVDepth, is proposed for accurate depth prediction for 3D object detection. We first study the working mechanism in existing 3D object detectors and reveal the unreliable depth in them. To address this, we introduce Camera-awareness Depth Prediction and Depth Refinement module with Explicit Depth Supervision in BEVDepth, making it able to generate robust depth prediction. BEVDepth obtains the capability to predict the trustworthy depth and obtains remarkable improvement compared to existing multi-view 3D detectors. Moreover, BEVDepth achieves the new state-of-the-art on nuScenes leaderboard with the help of Multi-frame Fusion schema and Efficient Voxel Pooling. We hope BEVDepth can serve as a strong baseline for future research in multi-view 3D object detection.

% \clearpage
\bibliography{aaai23}
\end{document}